\documentclass[letterpaper]{article} 
\usepackage[preprint]{aaai2027}
\usepackage[hyphens]{url}  
\usepackage{graphicx} 
\usepackage{natbib}  
\usepackage{caption} 
\usepackage{algorithm}
\usepackage{algorithmic}
\usepackage{amsmath}
\usepackage{times}
\usepackage{helvet}
\usepackage{graphicx}
\usepackage{multirow}
\usepackage{amsmath}
\usepackage{array}
\usepackage{amsmath,amssymb}
\usepackage{makecell}
\usepackage{courier}
\usepackage[switch]{lineno} 
\usepackage{newfloat}
\usepackage{listings}
\DeclareCaptionStyle{ruled}{labelfont=normalfont,labelsep=colon,strut=off} 
\floatstyle{ruled}
\newfloat{listing}{tb}{lst}{}
\floatname{listing}{Listing}

\usepackage{booktabs}

\title{OutLangSplat: 3D Language Gaussian Splatting for UAV Outdoor Scenes}
\author{
    Xia Yan\textsuperscript{\rm 1},
    He Wu\textsuperscript{\rm 1},
    Yanghui Xu\textsuperscript{\rm 1},
    Zizhao Wu\textsuperscript{\rm 2},
    Jiazhou Chen\textsuperscript{\rm 1}
    \thanks{Corresponding author: cjz@zjut.edu.cn}
}

\affiliations{
    \textsuperscript{\rm 1}Zhejiang University of Technology\\
    \textsuperscript{\rm 2}Hangzhou Dianzi University\\
}

\begin{document}

\maketitle

\begin{abstract}
3D Language Gaussian Splatting embeds open-vocabulary language features into 3D Gaussian Splatting, providing an efficient explicit representation for text-driven 3D scene understanding. However, existing methods are limited to indoor or small-scale scenes, and tend to fail in Unmanned Aerial Vehicle (UAV) outdoor scenes, where severe occlusions and long distance viewpoints often lead to incorrect semantic activations and missing target responses. In this paper, we present OutLangSplat which adapts language Gaussian representations to UAV outdoor scenes by improving feature representation and aggregation reliability. For the feature representation, a 2D-3D dual-branch representation with region-based alignment and fusion is designed to improve spatial consistency, reducing incomplete target responses and background misactivations. For the feature aggregation, we introduce a training-free contribution and consistency-aware Gaussian feature aggregation strategy that leverages pixel contribution reliability and cross-view semantic consistency to suppress unreliable responses from noisy viewpoints. A new dataset is provided by manually annotating various objects on four real-world public UAV outdoor scene datasets. To the best of our knowledge, it is the first accessible dataset of open-vocabulary 3D scene understanding for UAV outdoor scenes. Quantitative evaluations and ablation studies demonstrate that OutLangSplat outperforms SOTA methods on both open-vocabulary semantic segmentation and instance localization tasks. The datasets and codes will be open-sourced.
\end{abstract}

\section{Introduction}
\label{sec:intro}

Open-vocabulary 3D scene understanding aims to associate natural-language queries with corresponding spatial regions in 3D environments, enabling flexible recognition, localization, and interaction beyond predefined semantic categories~\cite{Choy20194DSC,Qi2016PointNetDL,Thomas2019KPConvFA}. This capability is particularly important for Unmanned Aerial Vehicle (UAV) outdoor scenes, where applications such as urban monitoring, land resource management, and emergency response require a fine-grained understanding of buildings, roads, vegetation, and other geographic objects. Compared with indoor scenes, UAV outdoor scenes typically cover larger spatial extents, contain more complex object layouts, and exhibit greater diversity in semantic granularities. In addition, severe occlusions, long-range observations, and cross-view variations from low-altitude aerial viewpoints make open-vocabulary 3D understanding in UAV outdoor scenes particularly challenging.

Recent advances in 3D Gaussian Splatting (3DGS)~\cite{Kerbl20233DGS} have enabled efficient and photorealistic representations of real-world 3D scenes, facilitating its widespread adoption across various 3D applications. By attaching or lifting language features to 3D Gaussians, existing methods have further extended 3DGS to support open-vocabulary querying of reconstructed scenes~\cite{Cheng2024OccamsLA,Qin2023LangSplat3L,Wu2024OpenGaussianTP,JunSeong2025DrSD}. However, most language-embedded 3DGS approaches are primarily designed for indoor environments. When directly applied to low-altitude UAV outdoor scenes, these methods are prone to incomplete target responses, blurred object boundaries, and semantic misactivations in irrelevant regions.

Existing methods predominantly rely on 2D vision-language features~\cite{Radford2021LearningTV} and directly aggregate multi-view features into 3D Gaussian representations~\cite{JunSeong2025DrSD,Marrie2024LUDVIGLU,Wang2025VisibilityAwareLA}. Although 2D vision-language models provide strong open-vocabulary semantic priors, their features lack explicit 3D structural constraints and are vulnerable to occlusions, viewpoint changes, and boundary mixing. Moreover, a pixel or region is typically rendered by multiple Gaussians, and ignoring their contribution differences may cause semantic mismatches at object boundaries and semantic boundary ambiguity. Existing aggregation methods also tend to treat all pixels and views as equally reliable, overlooking observation variations caused by occlusions, noises, and cross-view semantic inconsistency, thereby propagating unreliable semantic features into 3D Gaussian representations.

To address these challenges, we propose OutLangSplat, an open-vocabulary 3D scene understanding framework for UAV outdoor scenes. The framework includes the following key designs. (1) We use 2D semantic regions as alignment anchors and perform projection pooling and region-level fusion of the corresponding local 3D structural features. This improves the spatial consistency and completeness of semantic representations while reducing the semantic bias caused by relying solely on 2D features. (2) We assess feature reliability from pixel-level Gaussian contributions and view-level semantic consistency, and accordingly aggregate multi-view semantic features. This suppresses the propagation of unreliable features caused by occlusions, boundary mixing, noise, and cross-view semantic conflicts. Consequently, OutLangSplat produces more complete and robust language Gaussian representations. Furthermore, we construct a new open-vocabulary 3D dataset from four real-world publicly available UAV scenes, covering diverse semantic categories and varying semantic granularities. Experiments show that our method outperforms SOTA approaches on both localization and semantic segmentation tasks.

The main contributions of this work include:
\begin{itemize}
\item A 2D-3D dual-branch feature fusion framework with region-level alignment, which enhances the open-vocabulary semantic representation capability for UAV outdoor scenes.

\item A training-free pixel-level contribution and consistency-aware Gaussian feature aggregation strategy, which improves the robustness of multi-view feature aggregation.

\item The first open-vocabulary 3D scene understanding dataset for UAV real-world outdoor environments, filling the gap that existing evaluation benchmarks are mostly limited to indoor or small-scale scenes.

\end{itemize}

\section{Related Work}
\label{sec:related}

\subsection{Open-Vocabulary 3D Scene Representations}
Combining 3D representations with semantic information is essential for object localization, recognition, and semantic segmentation. Early studies primarily relied on NeRF~\cite{mildenhall2021nerf} based representations, where pretrained 2D vision-language features are distilled into neural radiance fields for text-driven querying and open-vocabulary 3D segmentation. Representative methods, including N3F~\cite{Tschernezki2022NeuralFF}, DFF~\cite{Kobayashi2022DecomposingNF}, LERF~\cite{Kerr2023LERFLE}, and OpenNeRF~\cite{Engelmann2024OpenNeRFOS}, demonstrated the potential of transferring 2D foundation-model knowledge to 3D representations. However, NeRF-based semantic fields require computationally expensive volumetric rendering and scene-specific optimization, limiting their efficiency and scalability in large-scale real-world scenes.

The emergence of 3D Gaussian Splatting (3DGS)~\cite{Kerbl20233DGS} has further advanced research on integrating language features into efficient and explicit 3D representations. Existing methods~\cite{Ye2023GaussianGS,Marrie2024LUDVIGLU,Zhai2025PanoGSGP} combine pretrained models such as CLIP~\cite{Radford2021LearningTV}, DINO~\cite{caron2021emerging}, SAM~\cite{Kirillov2023SegmentA}, and diffusion models with 3DGS to support semantic segmentation, language-driven localization, object retrieval, and scene editing. However, vision-language models typically produce high-dimensional semantic embeddings, and explicitly storing full language features for each Gaussian incurs substantial GPU memory and computational overhead. To alleviate this issue, prior methods~\cite{Qin2023LangSplat3L,Shi2023LanguageE3,Chen2024SLGaussianFL,Shorinwa2024FASTSplatFA} learn compact per-Gaussian semantic representations via autoencoder based compression, low-dimensional feature fields, quantization, or hash-based representations. More recently, Lang3D-XL~\cite{Krakovsky2025Lang3DXLLE} introduces low-dimensional semantic bottleneck features and a multi-resolution feature hash encoder for large-scale scenes, reducing the runtime and memory costs of language feature embedding.

Despite large progresses in language feature embedding, existing open-vocabulary 3D scene understanding methods are still mainly focused on indoor or small-scale scenes. Although Lang3D-XL~\cite{Krakovsky2025Lang3DXLLE} targets large-scale scenes, primarily focuses on buildings and their local structures, leaving the geographic objects in UAV outdoor scenes insufficiently explored. To this end, we extend open-vocabulary 3D scene understanding to UAV outdoor scenes and construct a corresponding dataset, providing a unified evaluation basis for semantic segmentation and multi-instance localization.

\subsection{Gaussian Feature Distillation and Aggregation}
To enable open-vocabulary understanding in 3D Gaussian scenes, recent studies have embedded 2D vision-language features into 3D Gaussian representations. LangSplat~\cite{Qin2023LangSplat3L} extracts multi-level CLIP features using SAM and compresses them with an autoencoder, whereas Feature3DGS~\cite{Zhou2023Feature3S} learns a mapping from low-dimensional Gaussian features to high-dimensional semantic embeddings. Other methods~\cite{Jiang2025VotesplatHV,Li2024InstanceGaussianAJ,Liang2024SuperGSegO3,Qu2024GOIF3,Piekenbrinck2025OpenSplat3DO3} first group Gaussian primitives into semantically meaningful clusters and then assign language embeddings to individual clusters. However, these approaches remain computationally expensive and rely on feature distillation with learnable language embeddings, which may propagate noises from 2D supervision into the learned 3D representations.
\begin{figure*}[t]
    \centering
    \includegraphics[
        width=1\textwidth,
        height=0.3\textheight]{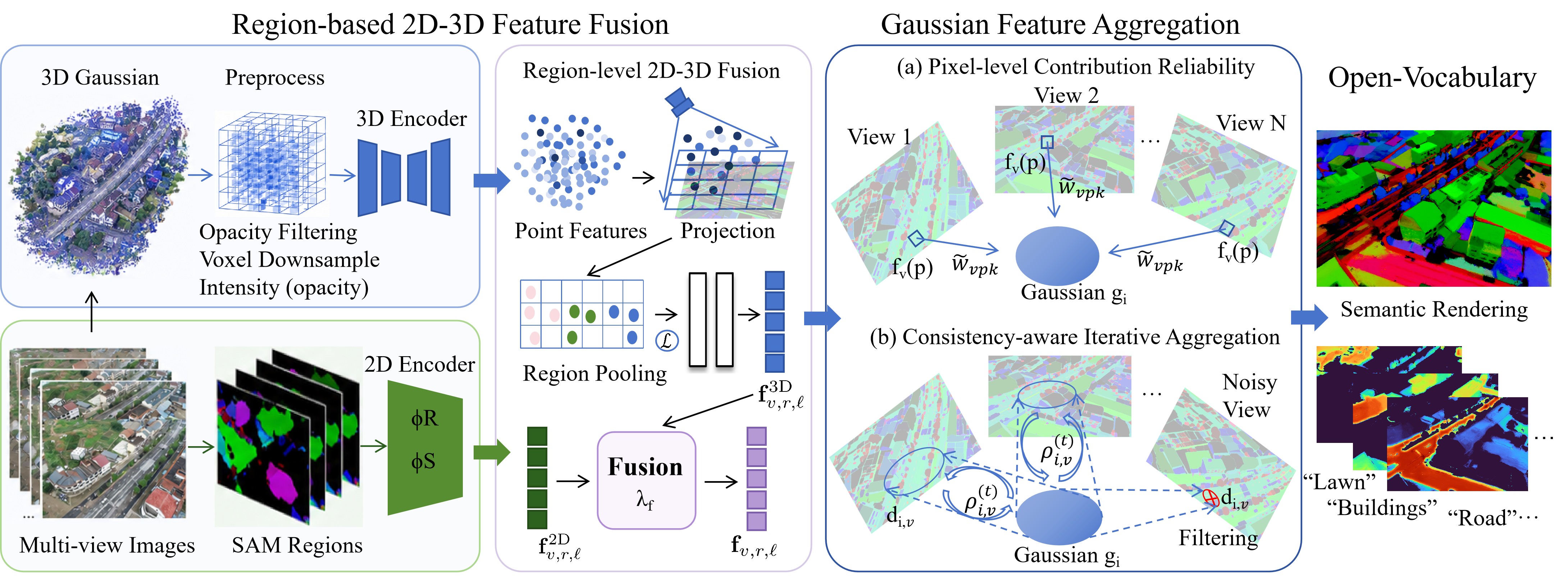}
    \caption{Overview of our OutLangSplat framework. Region-level 2D semantic and 3D structural features are fused, and then aggregated onto 3D Gaussians using pixel-level contribution reliability and cross-view semantic consistency, enabling open-vocabulary rendering, segmentation, and localization.}
    \label{fig:overview}
\end{figure*}

More recent studies explore more efficient feature aggregation strategies. Occam's LGS~\cite{Cheng2024OccamsLA} directly injects 2D language features into 3D Gaussians. Dr.Splat~\cite{JunSeong2025DrSD} employs Top-$K$ truncation to improve aggregation efficiency, but this strategy may discard many low-contribution Gaussians, weakening semantic responses and reducing foreground-background discriminability. LUDVIG~\cite{Marrie2024LUDVIGLU} aggregates 2D features according to multi-view rendering contributions and further incorporates 3D geometric information through graph diffusion. Although these methods avoid scene-specific semantic-field optimization, they remain vulnerable to noisy 2D observations, occlusions, and cross-view semantic inconsistency. VALA~\cite{Wang2025VisibilityAwareLA} introduces visibility-aware gating and robust aggregation to suppress low-quality observations and cross-view outliers. However, its contribution-based gating mechanism tends to filter out low-contribution yet valid surface Gaussians in UAV outdoor scenes, leading to incomplete semantic coverage and missing local responses. To address these limitations, we enhance truncated semantic representations with 3D structural features and design a training-free aggregation framework that improves robustness in UAV outdoor scenes by pixel-level contribution reliability and cross-view semantic consistency.
\section{Methodology}
\label{sec:method}

As illustrated in Fig.~\ref{fig:overview}, OutLangSplat consists of three main modules: 1) we construct region-based 2D-3D features. The 2D semantic branch extracts region-level open-vocabulary features from multi-view images using multimodal vision-language encoders, while the 3D structural branch extracts local structural features from the 3D Gaussian scene. The two feature types are then aligned and fused at the region level to form enhanced semantic representations. 2) We introduce a training-free robust Gaussian feature aggregation strategy that evaluates reliability using pixel-level Gaussian rendering contributions and improves cross-view semantic consistency through consistency-aware iterative aggregation. 3) The aggregated language Gaussian features support open-vocabulary semantic rendering, segmentation, and text-driven localization of geographic objects. We will review the 3D Gaussian representation and open-vocabulary querying as a background, and then introduce carefully region-based 2D-3D feature fusion, and the proposed Gaussian feature aggregation strategy in the remainder of this section.


\paragraph{3D Gaussian Splatting (3DGS)} represents a scene as a set of 3D Gaussian primitives~\cite{Kerbl20233DGS}. Each Gaussian \(G_i\) is parameterized by its center position \(\boldsymbol{\mu}_i\), covariance matrix \(\boldsymbol{\Sigma}_i\), opacity \(o_i\), and appearance information encoded by spherical harmonics \(\mathbf{c}_i\). The \(i\)-th Gaussian is defined as:
\begin{equation}
G_i(\mathbf{x})=\exp\left(-\frac{1}{2}(\mathbf{x}-\boldsymbol{\mu}_i)^\top
\boldsymbol{\Sigma}_i^{-1}(\mathbf{x}-\boldsymbol{\mu}_i)\right),
\end{equation}
During rendering, Gaussians are projected onto the image plane, and the color of a pixel \(\mathbf{v}\) is obtained by alpha blending:
\begin{equation}
\mathbf{C}(\mathbf{v})=
\sum_{i\in\mathcal{N}(\mathbf{v})}
\mathbf{c}_i(\mathbf{v})\alpha_i^{2\mathrm D}(\mathbf{v})
\prod_{j<i}\left(1-\alpha_j^{2\mathrm D}(\mathbf{v})\right),
\end{equation}
where \(\mathcal{N}(\mathbf{v})\) denotes the set of all Gaussians that cover \(\mathbf{v}\), and \(\alpha_i^{2\mathrm D}(\mathbf{v})\) is the effective opacity of the \(i\)-th Gaussian at \(\mathbf{v}\).

\paragraph{Open-Vocabulary Query.} The open-vocabulary semantic understanding task aims to generate semantic responses in a 3D scene according to arbitrary natural-language text queries \(q\). First, the text query \(q\) is mapped into the semantic feature space by the text encoder \(\phi_T(\cdot)\) of a vision-language model, producing a normalized text embedding: \(\mathbf{e}_q=
\frac{\phi_T(q)}{\|\phi_T(q)\|_2}
\), where \(\mathbf{e}_q\in\mathbb{R}^d\)and \(d\) denotes the feature dimension.
During semantic rendering, each Gaussian is assigned a semantic feature \(\mathbf{f}_i\in\mathbb{R}^d\). Following the alpha blending weights used in RGB rendering, the semantic feature of the pixel \(\mathbf{v}\) is:
\begin{equation}
\mathbf{F}(\mathbf{v})=
\frac{
\sum_{i\in\mathcal{N}(\mathbf{v})}
T_i(\mathbf{v})\alpha_i^{2\mathrm D}(\mathbf{v})\cdot \mathbf{f}_i
}{
\sum_{i\in\mathcal{N}(\mathbf{v})}
T_i(\mathbf{v})\alpha_i^{2\mathrm D}(\mathbf{v})+\epsilon
},
\label{eq:semantic_rendering}
\end{equation}
where \(T_i(\mathbf{v})=\prod_{j=1}^{i-1}\left(1-\alpha_j^{2\mathrm D}(\mathbf{v})\right)\) denotes the accumulated transmittance.
The semantic similarity between the pixel \(\mathbf{v}\) and text query \(q\) is computed by cosine similarity between the rendered feature and text embedding:
\begin{equation}
H_q(\mathbf{v})=
\frac{\mathbf{F}(\mathbf{v})^\top \mathbf{e}_q}
{\|\mathbf{F}(\mathbf{v})\|_2\|\mathbf{e}_q\|_2},
\label{eq:semantic_heatmap}
\end{equation}
Based on this heatmap \(H_q\), we further generate binary target regions and localization results for evaluating open-vocabulary 3D scene understanding performance.

\subsection{Region-based 2D-3D Feature Fusion}
UAV outdoor 3D scenes contain diverse geographic objects, posing challenges for open-vocabulary semantic understanding. Although 2D vision-language features provide strong semantic representations, they lack 3D spatial information and are susceptible to multi-view inconsistency and segmentation noise. 3D structural features offer better spatial consistency, but lack direct alignment with natural language queries. Existing 2D-3D fusion methods still suffer from mismatched representation granularities and feature spaces, making direct fusion ineffective in compensating for semantic response degradation caused by Top-\(K\) truncation. We therefore design a region-based 2D-3D dual-branch representation that integrates local 3D structural information into corresponding 2D semantic regions, enhancing the semantic discriminability and completeness of open-vocabulary features.

\paragraph{2D semantic feature branch.}
Geographic objects in UAV outdoor scenes exhibit substantial scale variations and diverse boundaries. Accordingly, we use multi-level semantic regions as the basic units for feature extraction and representation, preserving target region integrity while integrating complementary semantic information from different vision-language encoders. As shown in Fig.~\ref{fig:overview}, given an image \(I_v\) from view \(v\), we first partition it into multi-level regions, yielding
\(\mathcal{R}_v=\bigcup_{\ell=1}^{L}\{R_{v,r}^{\ell}\}_{r=1}^{M_v^{\ell}}\),
where \(R_{v,r}^{\ell}\) denotes the \(r\)-th semantic region at segmentation level \(\ell\) in view \(v\), and \(M_v^{\ell}\) is the number of regions at the level \(\ell\). For each region, we extract two complementary vision-language features and combine them through a weighted fusion:
\begin{equation}
\mathbf{f}_{v,r,\ell}^{2\mathrm D}
=
\lambda_{2\mathrm D}\phi_R(R_{v,r}^{\ell})
+
(1-\lambda_{2\mathrm D})\phi_S(R_{v,r}^{\ell}),
\label{eq:2d_feature_fusion}
\end{equation}
where \(\lambda_{2\mathrm D}\) controls the fusion of the two semantic sources, and \(\phi_R(\cdot)\) and \(\phi_S(\cdot)\) denote the corresponding vision-language encoders. By exploiting their complementary strengths in category recognition and geographic attribute representation, this design produces more complete region-level features for UAV outdoor scenes. In our implementation, RemoteCLIP~\cite{Liu2023RemoteCLIPAV} and RS5M~\cite{Zhang2023RS5MAG} are used as the two vision-language encoders.


\paragraph{3D structural feature branch.}
To introduce 3D geometric information, we export Gaussians from the optimized 3D Gaussian scene and filter unreliable Gaussians according to opacity. Let \({G}=\{g_i\}_{i=1}^{N}\) denote the set of 3D Gaussians, where each Gaussian has a center position \(\boldsymbol{\mu}_i\) and opacity \(o_i\). We retain Gaussians satisfying \(o_i>\tau_o\) and apply voxel downsampling to obtain a compact point set:
\(\mathcal{P}=\{(\mathbf{p}_k,\mathrm{id}_k)\}_{k=1}^{K}\)
where \(\mathbf{p}_k=\boldsymbol{\mu}_{\mathrm{id}_k}\)is the sampled 3D point coordinate, and \(\operatorname{id}_k\) denotes the index of its corresponding original Gaussian.
The downsampled point cloud is then converted into sparse convolutional input. For each point \(\mathbf{p}_k\), its voxel coordinate is computed as
\(
\mathbf{c}_k=\left\lfloor \frac{\mathbf{p}_k}{s} \right\rfloor,
\label{eq:voxel_coord}
\)
where \(s\) is the voxel size. 

We concatenate the normalized 3D coordinate with the corresponding Gaussian opacity to form the point feature:
\(
\mathbf{x}_k=[\widehat{\mathbf{p}}_k, o_{\operatorname{id}_k}]\in\mathbb{R}^{4}.
\label{eq:point_feature}
\)
The sparse input \(\mathcal{S}=\{(\mathbf{c}_k,\mathbf{x}_k)\}_{k=1}^{K}\) is then fed into a MinkUNet initialized with SegContrast-pretrained~\cite{Nunes2022SegContrast3P} weights. The encoder takes the sparse voxel coordinates and point features as input. After sparse convolutional encoding, it produces a 96-dimensional 3D structural feature for each sampled point:
\(
\{\mathbf{z}_k\}_{k=1}^{K}
=
\psi_{\mathrm{3D}}(\mathcal{S}),
\mathbf{z}_k\in\mathbb{R}^{96},
\)
where \(\psi_{3\mathrm D}(\cdot)\) denotes the 3D sparse convolutional encoder. \(\mathbf{z}_k\) mainly captures local geometric structure and spatial neighborhood relationships, providing a 3D structural prior for semantic fusion.

\paragraph{Region-level 2D-3D alignment and feature fusion.}
To align 3D structural features with 2D semantic regions, we project 3D points onto each 2D view and perform region-level pooling according to SAM~\cite{Kirillov2023SegmentA} masks. For the \(v\)-th view, let \(\Pi_v(\cdot)\) denote the camera projection function. The projected position of point \(\mathbf{p}_k\) on the image plane is given by
\(\mathbf{u}_{v,k}=\Pi_v(\mathbf{p}_k).\)
For each region \(R_{v,r}^{\ell}\), we define its valid projected point set as
\(\Omega_{v,r}^{\ell}=\{k \mid \Pi_v(\mathbf{p}_k)\in R_{v,r}^{\ell}\}.\)
The corresponding region-level 3D structural feature is obtained by average pooling:
\begin{equation}
\mathbf{f}_{v,r,\ell}^{3\mathrm D}
=
\frac{1}{\left|\Omega_{v,r}^{\ell}\right|}
\sum_{k\in\Omega_{v,r}^{\ell}}
\mathbf{z}_k,
\label{eq:3d_region_pooling}
\end{equation}

Since \(\mathbf{f}_{v,r,\ell}^{3\mathrm D}\in\mathbb{R}^{96}\) and the 2D semantic feature \(\mathbf{f}_{v,r,\ell}^{2\mathrm D}\in\mathbb{R}^{512}\) lie in different feature spaces, we learn a lightweight mapping function \(h_\theta(\cdot)\) to distill the 3D structural feature into the vision-language feature space:
\(\widetilde{\mathbf{f}}_{v,r,\ell}^{3\mathrm D}
=
h_\theta(\mathbf{f}_{v,r,\ell}^{3\mathrm D})
\in\mathbb{R}^{512}.\)
Here, \(h_\theta(\cdot)\) is implemented as a two-layer MLP and trained using the region-level 2D semantic feature as the teacher signal. The distillation objective is defined as a cosine loss:
\(\mathcal{L}_{\mathrm{distill}}
=
1-
\cos\left(
h_\theta(\mathbf{f}_{v,r,\ell}^{3\mathrm D}),
\mathbf{f}_{v,r,\ell}^{2\mathrm D}
\right).\)

Finally, we fuse the original 2D open-vocabulary semantic feature with the mapped 3D structure enhanced feature:
\begin{equation}
\mathbf{f}_{v,r,\ell}
=
\lambda_f\mathbf{f}_{v,r,\ell}^{2\mathrm D}
+
(1-\lambda_f)\widetilde{\mathbf{f}}_{v,r,\ell}^{3\mathrm D},
\label{eq:2d_3d_fusion}
\end{equation}
where \(\lambda_f\) controls the fusion ratio between the 2D semantic feature and the 3D structure enhanced feature. The fused feature \(\mathbf{f}_{v,r,\ell}\) provides a more reliable semantic input for subsequent Gaussian aggregation and text-driven querying.

\subsection{Consistency-aware Gaussian Feature Aggregation}
Feature aggregation integrates multi-view features into the corresponding 3D Gaussians to construct semantically expressive Gaussian representations. Although the fused region feature \(\mathbf{f}_{v,r,\ell}\) contains both 2D semantics and 3D structural cues, conventional average aggregation~\cite{Marrie2024LUDVIGLU} ignores reliability variations across pixels and views. VALA~\cite{Wang2025VisibilityAwareLA} improves robustness through visibility-aware gating but suppresses low-contribution yet valid surface Gaussians, causing semantic loss. We therefore propose a training-free aggregation method that evaluates observation reliability using pixel-level Gaussian contributions and multi-view semantic consistency.

\paragraph{Pixel-level contribution reliability.}
As shown in Fig.~\ref{fig:overview}(a), for each pixel \(p\) in view \(v\), we rank the Gaussians whose projected footprints cover the pixel in descending order of their alpha-blending contributions to form an adaptive Top-\(K\) set, where \(K\) ranges from 16 to 32, our selection stops once 90\% of the total contribution is covered.
Let \(\operatorname{id}_{v,p,k}\) denote the index of the \(k\)-th candidate Gaussian, and \(w_{v,p,k}\) denote its alpha blending weight, normalize the Top-\(K\) weights within each pixel: \(\pi_{v,p,k}={w_{v,p,k}}/({\sum_{k=1}^{K} w_{v,p,k}+\epsilon}).\)

\vspace{0.1cm}

\begin{figure}[t]
    \centering
    \includegraphics[width=0.95\columnwidth]{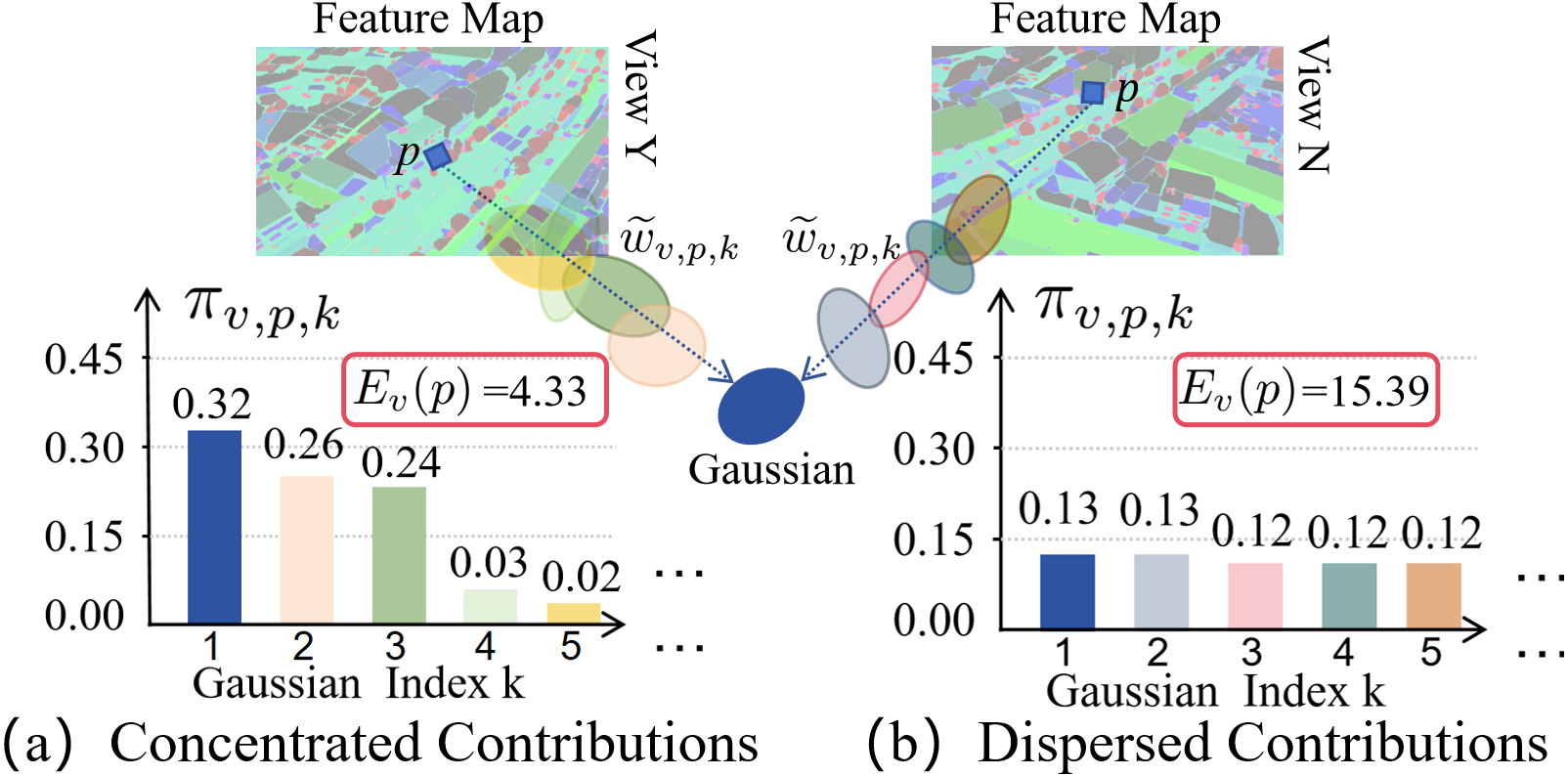}
    \caption{
    Pixel reliability measured by the inverse Simpson index $E_v(p)$: concentrated contributions indicate high reliability, while dispersed contributions indicate low reliability.
    }
    \label{fig:pixel-reliability}
\end{figure}

As illustrated in Fig.~\ref{fig:pixel-reliability}, after Top-\(K\) selection, Gaussian contribution distributions vary across pixels: contributions are concentrated on a few Gaussians in visible regions but become more dispersed in multi-layer gaussian mixtures, weakening pixel-Gaussian correspondences. Therefore, we use the inverse Simpson index to quantify contribution dispersion and assess the aggregation reliability of each pixel:
\begin{equation}
E_v(p)
=
\frac{1}
{\sum_{k=1}^{K}\pi_{v,p,k}^{2}+\epsilon}.
\label{eq:effective_gaussian_number}
\end{equation}

\noindent where \(E_v(p)\) can be interpreted as the effective number of contributing Gaussians. A small \(E_v(p)\) indicates that the pixel is mainly explained by a small number of dominant Gaussians, leading to a clearer semantic assignment. In contrast, a large \(E_v(p)\) suggests dispersed contributions, which often occur near boundaries, occlusions, or mixed regions.

To suppress unreliable pixel features, we define the pixel quality weight as
\(q_v(p)=q_{\mathrm{sum}}(p)\cdot q_{\mathrm{eff}}(p),\)
where
\(q_{\mathrm{sum}}(p)
=
\operatorname{clip}
(
\sum_{k=1}^{K} w_{v,p,k},
q_{\min},
1
),
\label{eq:q_sum}\)
measures how completely the selected Top-\(K\) Gaussians explain the pixel, and
\(q_{\mathrm{eff}}(p)
=
\exp \left(-\max(E_v(p)-\tau_E,0)/{\kappa} \right)
\)
penalizes pixels with overly dispersed contributions. \(\tau_E\) is the threshold for the effective number of Gaussians, and \(\kappa\) controls the decay strength. We set \(\tau_E=12\) and \(\kappa=8\) in all of our experiments.The final effective semantic contribution weight from pixel \(p\) to Gaussian \(\operatorname{id}_{v,p,k}\) is defined as

\begin{equation}
\widetilde{w}_{v,p,k}
=
w_{v,p,k}\cdot M_v(p)\cdot
q_{\mathrm{sum}}(p)\cdot q_{\mathrm{eff}}(p).
\end{equation}

\noindent where \(M_v(p)\) denotes the valid mask of semantic features.
With the pixel-level quality weights, we aggregate all pixel observations within each view into Gaussian-level observations. For Gaussian \(g_i\), its observation feature in view \(v\) is:
\begin{equation}
\mathbf{f}_{i,v}
=
\frac{
\sum_{p,k}
\mathbb{I}[\operatorname{id}_{v,p,k}=i] \cdot
\widetilde{w}_{v,p,k} \cdot \mathbf{f}_v(p)
}{
\sum_{p,k}
\mathbb{I}[\operatorname{id}_{v,p,k}=i] \cdot
\widetilde{w}_{v,p,k}
+\epsilon
},
\label{eq:view_gaussian_feature}
\end{equation}
where \(\mathbf{f}_v(p)\) denotes the fused semantic feature assigned to pixel \(p\). The corresponding visibility, or observation strength, is defined as
\(d_{i,v}
=
\sum_{p,k}
\mathbb{I}[\operatorname{id}_{v,p,k}=i] \cdot
\widetilde{w}_{v,p,k}.\)

When the visibility \(d_{i,v}\) is too small, the Gaussian is only weakly observed in the current view and is likely to be affected by noise. Therefore, we retain only reliable view-level observations:
\(\mathcal{O}_i
=
\{(\mathbf{f}_{i,v},d_{i,v})\mid d_{i,v}>\tau_d\}.\)


\paragraph{Consistency-aware iterative aggregation.}
As shown in Fig.~\ref{fig:overview}(b), we further introduce an iterative view-level semantic consistency refinement. For Gaussian \(g_i\), we first compute its initial semantic feature by weighted averaging all valid view-level observations:
\begin{equation}
\mathbf{F}_i^{(0)}
=
\frac{
\sum_{v\in\mathcal{V}_i} d_{i,v}\cdot \mathbf{f}_{i,v}
}{
\sum_{v\in\mathcal{V}_i} d_{i,v}+\epsilon
},
\label{eq:initial_gaussian_feature}
\end{equation}
where \(\mathcal{V}_i\) denotes the set of valid views for Gaussian \(g_i\).

At iteration \(t\), we compute the cosine consistency between the view-level observation \(\mathbf{f}_{i,v}\) and the aggregated feature from the previous iteration:
\(c_{i,v}^{(t)}
=
\cos(\mathbf{f}_{i,v},\mathbf{F}_i^{(t-1)}).\)

Based on this consistency score, the robust view-level weight is defined as
\(\rho_{i,v}^{(t)}
=
d_{i,v}\cdot
\exp(\beta_t c_{i,v}^{(t)}),\)
where \(\beta_t\) is a sharpening coefficient. Observations consistent with the current aggregated feature are assigned larger weights, while inconsistent observations are automatically down-weighted. The Gaussian semantic feature is then updated as
\begin{equation}
\mathbf{F}_i^{(t)}
=
\frac{
\sum_{v\in\mathcal{V}_i}\rho_{i,v}^{(t)}\cdot \mathbf{f}_{i,v}
}{
\sum_{v\in\mathcal{V}_i}\rho_{i,v}^{(t)}+\epsilon
}.
\label{eq:robust_feature_update}
\end{equation}
After \(T\) iterations, the final semantic feature of Gaussian \(g_i\) is obtained as
\(\mathbf{f}_i=\mathbf{F}_i^{(T)},\)
which is stored as the language feature of the Gaussian for subsequent semantic rendering.

Finally, for a text query \(q\), Eqs.~\ref{eq:semantic_rendering} and~\ref{eq:semantic_heatmap} render the aggregated Gaussian features and compute similarity with \(\mathbf{e}_q\), producing semantic heatmaps for open-vocabulary recognition and localization.

\section{Experiments and Discussion}

\begin{table}[b]
\centering
\caption{Statistics of the proposed UAV outdoor open-vocabulary 3D scene understanding datasets.}
\label{tab:dataset}

\setlength{\tabcolsep}{2pt} 
\renewcommand{\arraystretch}{1} 
\footnotesize
\begin{tabular}{lccc>{\centering\arraybackslash}p{1.cm}}
\hline
\textbf{Scenes} &
\textbf{Area ($\mathrm{m}^2$)} &
\textbf{Height ($\mathrm{m}$)} &
\textbf{Gaussians} &
\textbf{Regions} \\
\hline
Buildings1 & 22{,}000 & 64.1 & 2.02M & \#204 \\
Buildings2 & 21{,}000 & 67.5 & 4.65M & \#197 \\
Polytech   & 15{,}000 & 113.4 & 3.53M & \#150 \\
Campus     & 56{,}000 & 119.3 & 3.77M & \#109 \\
\hline
\end{tabular}
\end{table}

\label{sec:experiment}

\subsection{Experimental Setup}
\begin{figure*}[t]
    \centering
    \includegraphics[width=\textwidth]{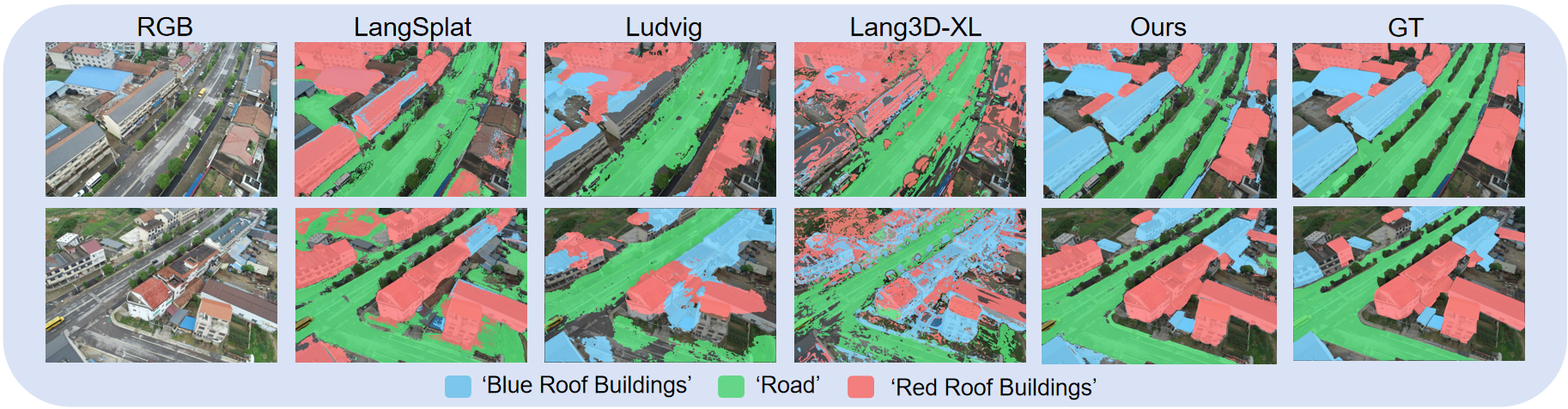}
    \caption{Qualitative comparison of open-vocabulary semantic segmentation. Our method produces more complete regions, clearer boundaries, and fewer false activations.
}
    \label{fig:compare-visual} 
\end{figure*}

\begin{figure}[t]
    \centering
    \setlength{\tabcolsep}{1pt}
    \renewcommand{\arraystretch}{0.8}
    \scriptsize

    \begin{tabular}{c c c c c}
        &
        \textbf{LangSplat} &
        \textbf{LUDVIG} &
        \textbf{Lang3D-XL} &
        \textbf{Ours} \\

        \raisebox{-0.1\height}{\rotatebox{90}{\textit{White Buildings}}} &
        \includegraphics[width=0.22\columnwidth]{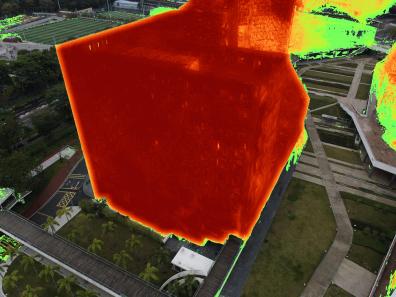} &
        \includegraphics[width=0.22\columnwidth]{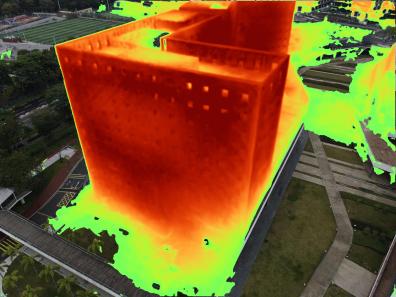} &
        \includegraphics[width=0.22\columnwidth]{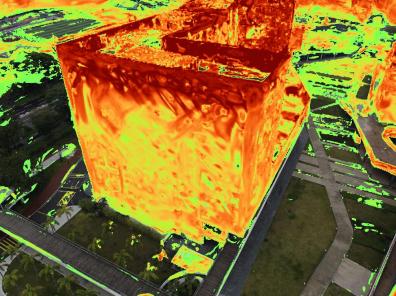} &
        \includegraphics[width=0.22\columnwidth]{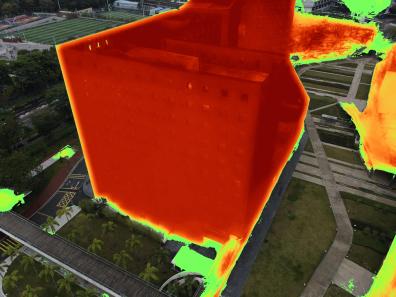} \\

        \raisebox{0.5\height}{\rotatebox{90}{\textit{Lawn}}} &
        \includegraphics[width=0.22\columnwidth]{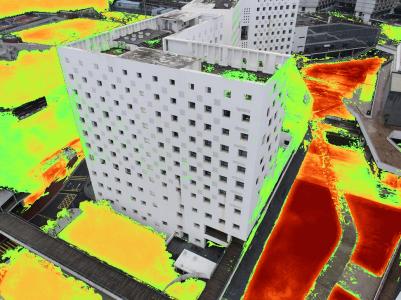} &
        \includegraphics[width=0.22\columnwidth]{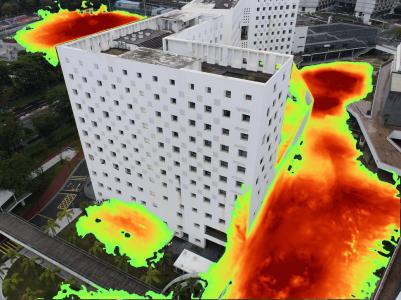} &
        \includegraphics[width=0.22\columnwidth]{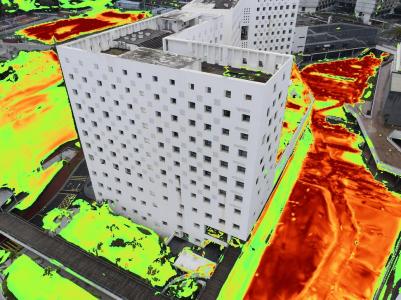} &
        \includegraphics[width=0.22\columnwidth]{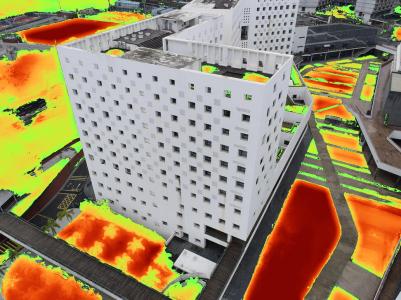} \\

        \raisebox{0.2\height}{\rotatebox{90}{\textit{Buildings}}} &
        \includegraphics[width=0.22\columnwidth]{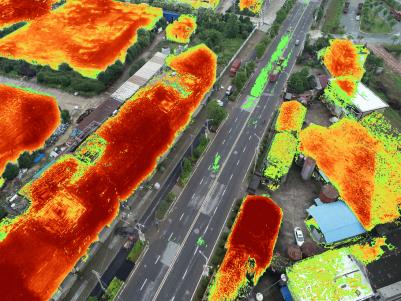} &
        \includegraphics[width=0.22\columnwidth]{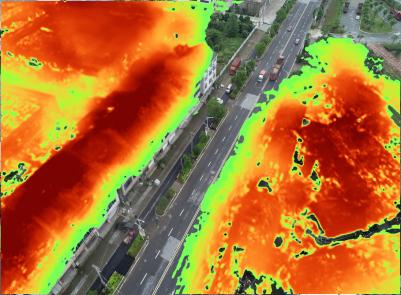} &
        \includegraphics[width=0.22\columnwidth]{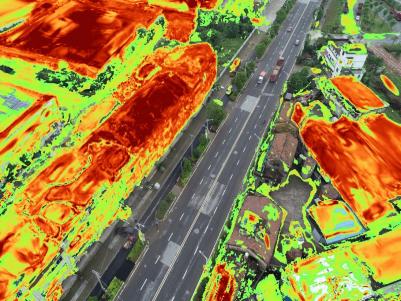} &
        \includegraphics[width=0.22\columnwidth]{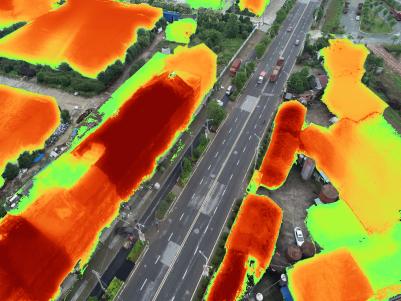} \\
    \end{tabular}

    \caption{
    Qualitative comparison on open-vocabulary queries.
    Our method produces more accurate target localization and cleaner object boundaries.
    }
    \label{fig:compare-qual}
\end{figure}
\paragraph{Dataset.}
To evaluate text-driven understanding of geographic objects in real-world UAV outdoor 3D scenes, we construct an open-vocabulary 3D scene understanding dataset. As shown in Table~\ref{tab:dataset}, it comprises four public real-world scenes from InstanceBuilding (Buildings1 and 2)~\cite{Chen20213DIS}  and UrbanScene3D (Polytech and Campus)~\cite{Lin2021CapturingRA}. Reconstructed from real UAV imagery, these scenes vary substantially in spatial coverage, building layout, object density, and occlusions, and contain diverse geographic objects such as buildings, roads, lawns, and trees. In total, the dataset includes 660 annotated target regions.
Using a self-developed annotation script, we manually delineate the visible regions of each target across multiple views with pixel-level polygons and assign corresponding attribute labels, producing annotations for semantic segmentation and object localization. The annotation process took approximately one month and followed consistent labeling and boundary criteria to ensure annotation quality.

\paragraph{Evaluation metrics.}
We use mIoU and mAcc to evaluate region overlap and target recognition accuracy. For large and spatially distributed objects in UAV outdoor scenes, we further report Loc@40, Loc@50, and Loc@60, where an instance is correctly localized if the predicted region covers more than the specified proportion of its annotated area.

\paragraph{Evaluation on datasets.}
We compare our method with three representative approaches: LangSplat~\cite{Qin2023LangSplat3L}, Lang3D-XL~\cite{Krakovsky2025Lang3DXLLE}, and LUDVIG~\cite{Marrie2024LUDVIGLU}, representing 2D feature lifting, large-scale language-embedded 3D Gaussian Splatting, and training-free feature lifting. All experiments run on NVIDIA 3090, more results are in the supplementary material.

\begin{table}[t]
\centering
\caption{Localization accuracy comparison under coverage thresholds of 40\% / 50\% / 60\%.}
\label{tab:loc_compare}
\scriptsize
\setlength{\tabcolsep}{2pt}
\renewcommand{\arraystretch}{1.05}
\resizebox{\columnwidth}{!}{
\begin{tabular}{lcccc}
\hline
\textbf{Scenes} &
\textbf{LangSplat} &
\textbf{LUDVIG} &
\textbf{Lang3D-XL} &
\textbf{Ours} \\
\hline
Buildings1 & 85.3/83.8/80.3 & 61.3/59.8/57.8 & 56.4/46.6/38.2 & \textbf{93.6}/\textbf{91.7}\textbf{/86.3} \\
Buildings2 & 86.2/84.7/83.2 & 72.1/69.5/65.4 & 53.8/45.2/39.1 & \textbf{89.8}/\textbf{86.8}/\textbf{85.2} \\
Polytech   & 92.0/89.3/86.6 & 73.3/72.0/69.3 & 76.0/69.3/68.0 & \textbf{93.3}/\textbf{91.3}/\textbf{89.3} \\
Campus     & \textbf{87.1}/84.4/83.4 & 69.7/66.9/62.3 & 80.7/80.7/76.1 & \textbf{87.1}/\textbf{85.3}/\textbf{84.4} \\
\hline
Overall    & 87.7/{85.6}/83.4 & 69.1/{67.1}/63.7 & 66.7/{60.5}/55.4 & \textbf{90.9}/\textbf{88.8}/\textbf{86.3} \\
\hline
\end{tabular}
}
\end{table}

\begin{table}[t]
\centering
\footnotesize
\caption{Quantitative comparison on 3D semantic segmentation over the proposed dataset with mIoU (\%) \& mAcc (\%).}
\label{tab:seg_compare}
\setlength{\tabcolsep}{3pt}
\renewcommand{\arraystretch}{0.95}
\begin{tabular}{lcccccccc}
\hline
\multirow{2}{*}{\textbf{Scenes}}
& \multicolumn{2}{c}{\textbf{LangSplat}}
& \multicolumn{2}{c}{\textbf{Ludvig}}
& \multicolumn{2}{c}{\textbf{Lang3D-XL}}
& \multicolumn{2}{c}{\textbf{Ours}} \\
\cline{2-9}
& {\scriptsize mIoU} & {\scriptsize mAcc}
& {\scriptsize mIoU} & {\scriptsize mAcc}
& {\scriptsize mIoU} & {\scriptsize mAcc}
& {\scriptsize mIoU} & {\scriptsize mAcc} \\
\hline
Buildings1 & 45.7 & 82.4 & 43.9 & 72.9 & 41.3 & 64.5 & \textbf{68.3} & \textbf{86.6} \\
Buildings2 & 51.4 & 84.8 & 38.9 & 80.9 & 41.1 & 64.7 & \textbf{72.9} & \textbf{88.1} \\
Polytech & 70.3 & \textbf{96.2} & 47.1 & 69.2 & 54.3 & 76.1 & \textbf{82.9} & 93.9 \\
Campus & 54.8 & 82.2 & 49.7 & 72.4 & 58.9 & 82.0 & \textbf{72.4} & \textbf{85.9} \\
\hline
\end{tabular}
\end{table}

\begin{table}[b]
\centering
\caption{Category-level 3D semantic segmentation comparison on selected open-vocabulary queries with IoU (\%).}
\label{tab:category_compare}
\scriptsize
\setlength{\tabcolsep}{3pt}
\renewcommand{\arraystretch}{1.05}

\begin{tabular}{lccccccc}
\hline
\textbf{Methods} &
\textbf{Tree} &
\textbf{Road} &
\textbf{Lawn} &
\textbf{Buildings} &
\makecell{\textbf{White}\\\textbf{Buildings}} &
\makecell{\textbf{Red Roof}\\\textbf{Buildings}} &
\makecell{\textbf{Blue Roof}\\\textbf{Buildings}} \\
\hline
LangSplat & 47.5 & 44.1 & 59.4 & 77.2 & 79.7 & 42.6 & 26.4 \\
LUDVIG & 40.1 & 38.6 & 33.3 & 69.4 & 65.2 & 42.4 & 32.8 \\
Lang3D-XL & 51.8 & 52.3 & 40.1 & 64.5 & 60.1 & 30.9 & 24.3 \\
Ours & \textbf{69.2} & \textbf{67.3} & \textbf{68.1} & \textbf{85.5} & \textbf{91.5} & \textbf{78.1} & \textbf{59.3} \\
\hline
\end{tabular}
\end{table}

\paragraph{Quantitative and Qualitative Comparison.}
As shown in Figs.~\ref{fig:compare-visual} and~\ref{fig:compare-qual}, LangSplat suffers from boundary leakage and false background activations, while LUDVIG produces incomplete target regions and local confusion. Lang3D-XL exhibits substantial noise. In contrast, our method produces more complete target regions and clearer boundaries while effectively suppressing non-target activations.
As shown in Table~\ref{tab:loc_compare}, our method achieves a Loc@50 of \(88.8\%\) and consistently outperforms competing methods across Loc@\{40, 50, 60\}, demonstrating its ability to localize multiple spatially distributed instances in outdoor scenes. Table~\ref{tab:seg_compare} show that our method achieves the best overall mIoU and mAcc, reaching \(82.9\%\) mIoU and \(93.9\%\) mAcc on Polytech. It also obtains the highest IoU across selected open-vocabulary queries in Table~\ref{tab:category_compare}, with a peak IoU of \(91.5\%\) for `white buildings'. 

Moreover, our training-free aggregation strategy completes feature aggregation achieving runtime comparable to LUDVIG while avoiding the costly optimization required by LangSplat and Lang3D-XL. This substantially reduces processing time and computational overhead while maintaining superior localization and segmentation performance, which makes our method more suitable for UAV outdoor scenes.

\subsection{Ablation Studies}

We conduct ablation studies on the region-based 2D-3D feature fusion and consistency-aware Gaussian feature aggregation modules to evaluate their effectiveness.

\paragraph{Ablation on feature fusion.}
Figure~\ref{fig:ablation-fusion} compares different feature configurations for the queries ``buildings'' and ``road''. RemoteCLIP alone provides more stable semantic responses for buildings, whereas RS5M responds more strongly to structurally salient regions such as roads and vegetation. Weighted 2D fusion combines their complementary semantic cues and reduces local response missing. However, without 3D region fusion, Top-\(K\) aggregation still weakens responses in some target regions, resulting in lower activations and reduced spatial continuity.By further incorporating 3D structural features, the proposed 2D-3D region fusion produces stronger, more concentrated, and spatially continuous target responses while suppressing noisy activations in non-target regions. Table~\ref{tab:feature_ablation} show that 2D-3D fusion achieves the best mIoU and mAcc, confirming the benefit of 3D structural information for open-vocabulary geographic object understanding. We further evaluate \(\lambda_{2D}\in\{0.5,0.6,0.7\}\) and \(\lambda_f\in\{0.6,0.7,0.8\}\). As shown in Table~\ref{tab:fusion_param}, the best overall performance is obtained when \(\lambda_{2D}=0.6\) and \(\lambda_f=0.7\).

\begin{table}[b]
\centering
\caption{Ablation study on feature fusion parameters $\lambda_{2D}$ and $\lambda_f$. We report mIoU, mAcc, and mLoc@50 (\%).}
\label{tab:fusion_param}

\setlength{\tabcolsep}{3pt}
\renewcommand{\arraystretch}{1.05}
\footnotesize

\begin{tabular}{ccccc}
\hline
$\lambda_{2D}$ & $\lambda_f$ & {mIoU} \(\uparrow\) & {mAcc} \(\uparrow\) & {mLoc@50} \(\uparrow\) \\
\hline
0.5 & 0.6/0.7/0.8 & 64.9/67.5/63.7 & 84.9/85.2/85.2 & 82.8/88.2/86.2 \\
0.6 & 0.6/0.7/0.8 & 66.2/\textbf{68.3}/62.4 & \textbf{86.8}/86.6/81.6 & 89.7/\textbf{91.6}/86.2 \\
0.7 & 0.6/0.7/0.8 & 64.6/64.1/62.3 & 80.5/82.4/86.5 & 86.7/87.2/89.2 \\
\hline
\end{tabular}
\end{table}

\begin{table}[h]
\centering
\caption{Comparison of different Gaussian feature aggregation strategies with mIoU (\%) \& mAcc (\%).}
\label{tab:aggregation_ablation}
\footnotesize
\setlength{\tabcolsep}{7pt}
\renewcommand{\arraystretch}{1.10}
\begin{tabular}{lccc}
\toprule
\textbf{Metrics} &
\makecell{\textbf{Projection}\textbf{ Agg.}} &
\makecell{\textbf{VALA}\textbf{ Agg.}} &
\textbf{Our Agg.} \\
\midrule
mIoU \(\uparrow\) & 65.44 & 62.87 & \textbf{68.31} \\
mAcc \(\uparrow\) & 84.00 & 82.04 & \textbf{86.61} \\
\bottomrule
\end{tabular}
\end{table}

\begin{table}[t]
\centering
\caption{Ablation study of different feature configurations on semantic segmentation with mIoU (\%) \& mAcc (\%).}
\label{tab:feature_ablation}
\footnotesize
\setlength{\tabcolsep}{2.8pt}
\renewcommand{\arraystretch}{1.08}
\begin{tabular}{lcccc}
\toprule
\textbf{Metrics} &
\makecell{\textbf{RemoteCLIP}} &
\makecell{\textbf{RS5M}} &
\makecell{\textbf{w/o}\textbf{3D fusion}} &
\makecell{\textbf{2D-3D} \textbf{fusion}} \\
\midrule
mIoU \(\uparrow\) & 49.91 & 58.99 & 63.90 & \textbf{68.31} \\
mAcc \(\uparrow\) & 66.89 & 78.69 & 77.41 & \textbf{86.61} \\
\bottomrule
\end{tabular}
\end{table}

\begin{figure}[t]
    \centering
    \setlength{\tabcolsep}{1pt}
    \renewcommand{\arraystretch}{0.8}
    \scriptsize

    \begin{tabular}{ccccc}
        \textbf{RemoteCLIP} &
        \textbf{RS5M} &
        \textbf{w/o 3D Fusion} &
        \textbf{2D--3D Fusion} &
        \textbf{GT} \\

        
        \includegraphics[width=0.19\columnwidth]{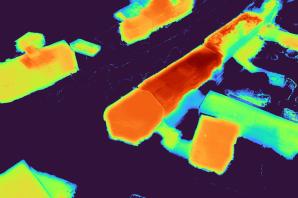} &
        \includegraphics[width=0.19\columnwidth]{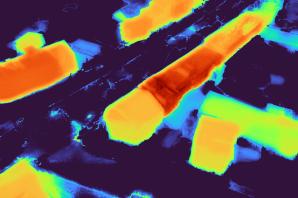} &
        \includegraphics[width=0.19\columnwidth]{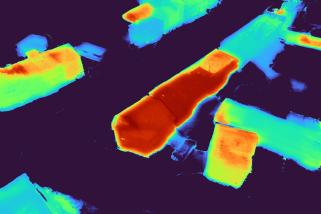} &
        \includegraphics[width=0.19\columnwidth]{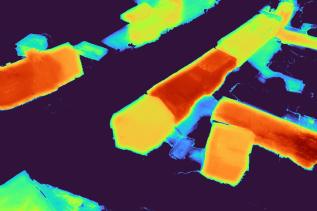} &
        \includegraphics[width=0.19\columnwidth]{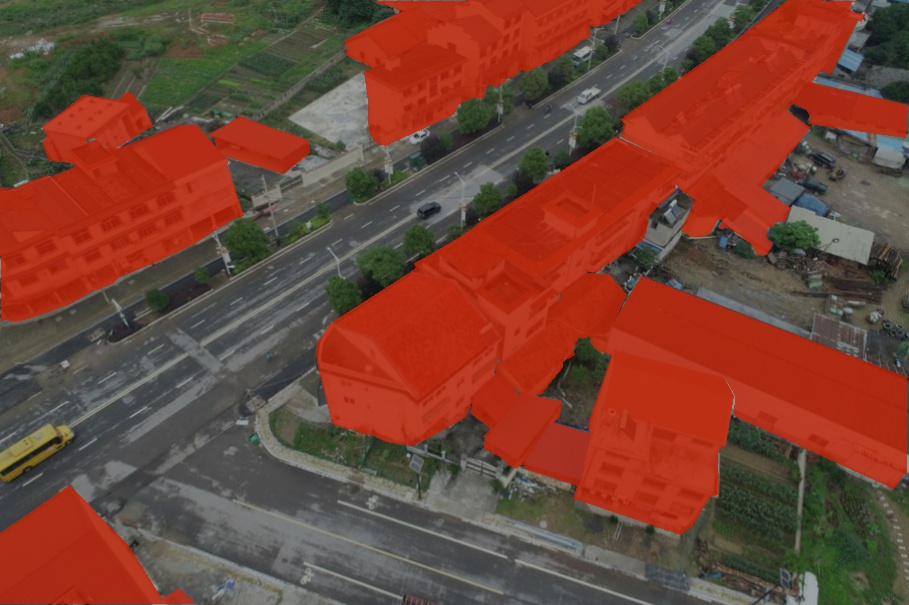} \\

        \includegraphics[width=0.19\columnwidth]{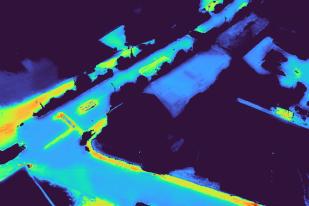} &
        \includegraphics[width=0.19\columnwidth]{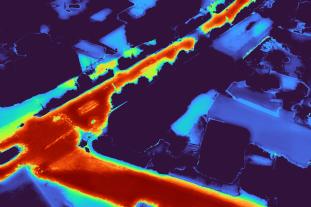} &
        \includegraphics[width=0.19\columnwidth]{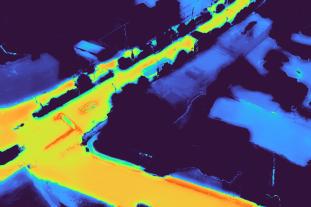} &
        \includegraphics[width=0.19\columnwidth]{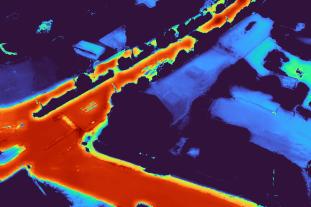} &
        \includegraphics[width=0.19\columnwidth]{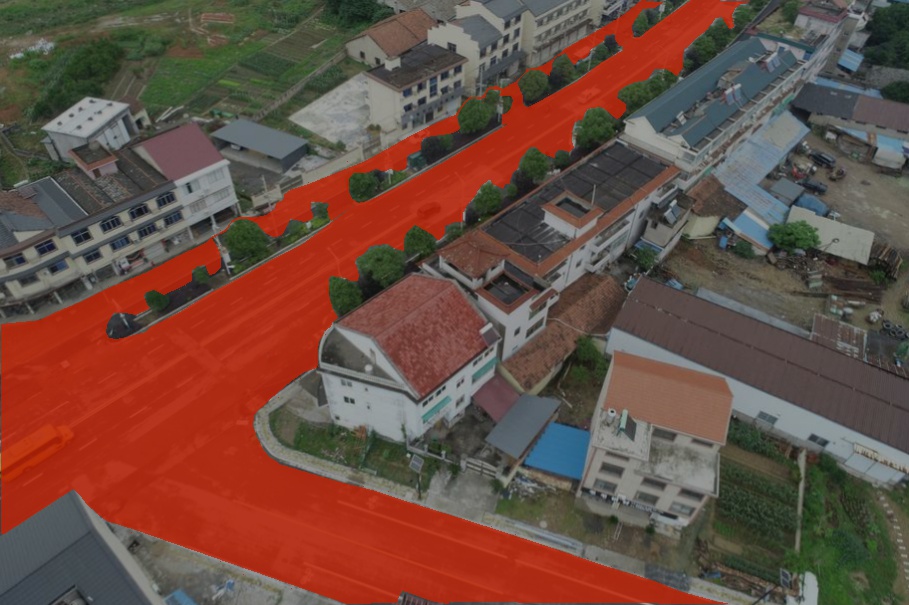} \\
    \end{tabular}

    \caption{
    Ablation study on feature fusion for the queries ``buildings'' and ``road.'' 
    From left to right: ground truth, RemoteCLIP only, RS5M only, without 3D region fusion, and the proposed 2D--3D fusion.
    }
    \label{fig:ablation-fusion}
\end{figure}

\begin{figure}[H]
    \centering
    \setlength{\tabcolsep}{1pt}
    \renewcommand{\arraystretch}{0.95}
    \scriptsize

    \begin{tabular}{ccc}
        \includegraphics[width=0.32\columnwidth]{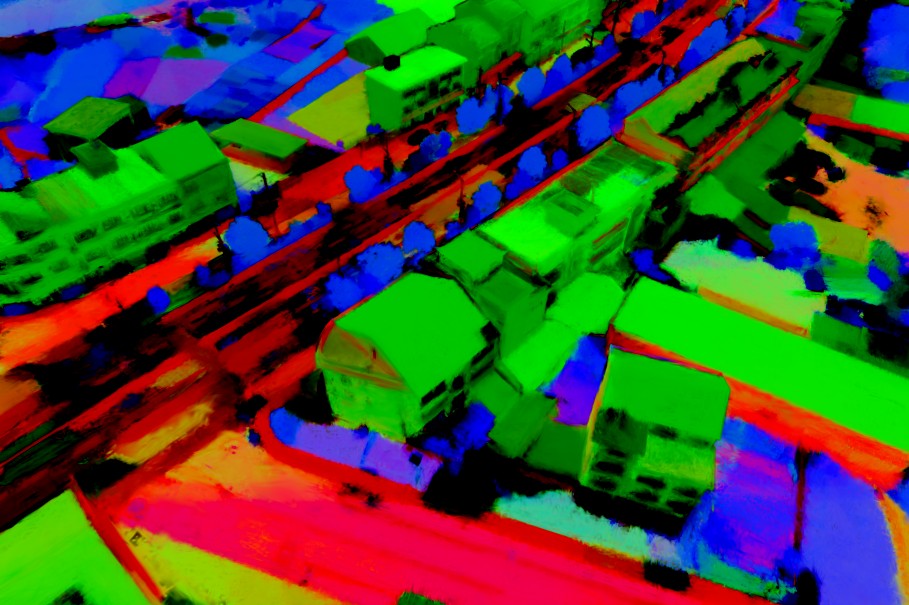} &
        \includegraphics[width=0.32\columnwidth]{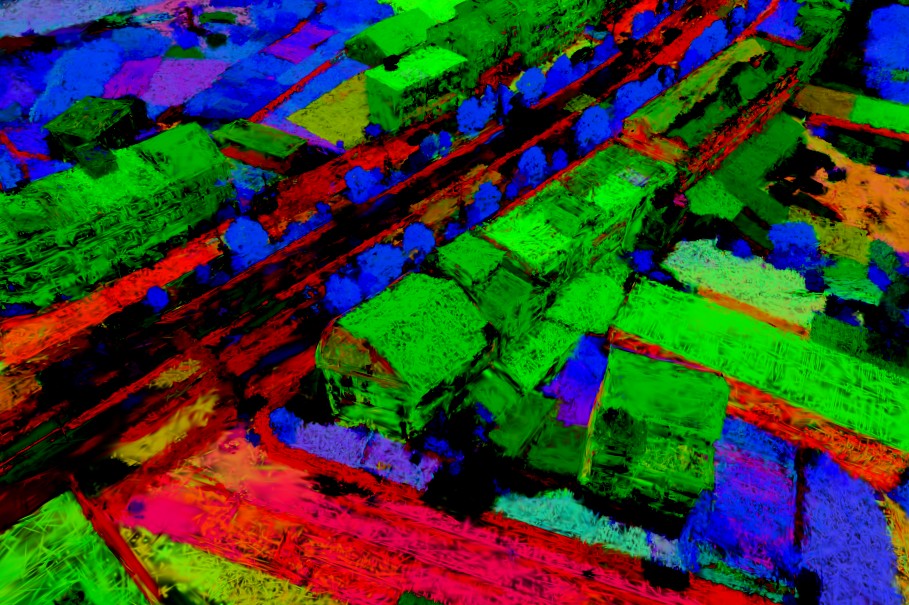} &
        \includegraphics[width=0.32\columnwidth]{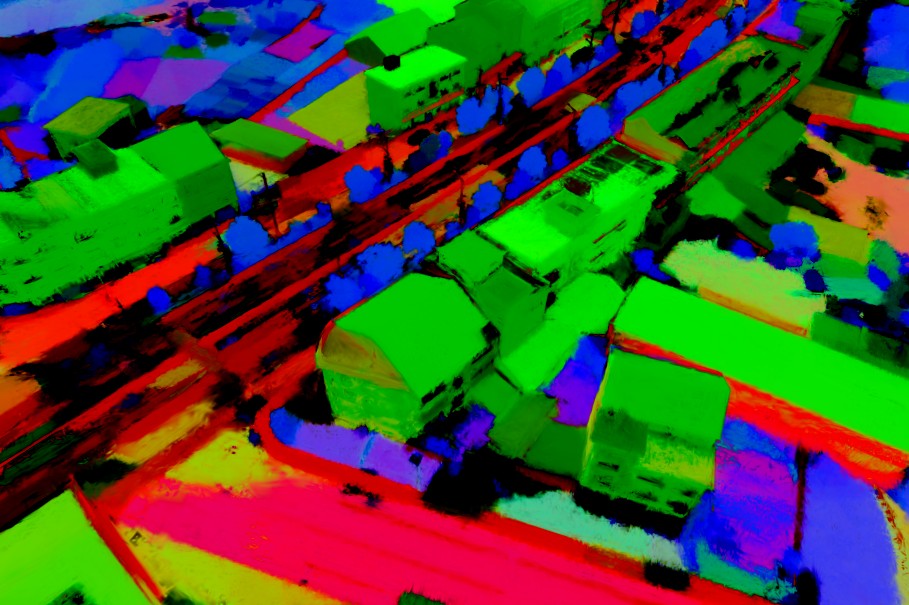} \\

        \includegraphics[width=0.32\columnwidth]{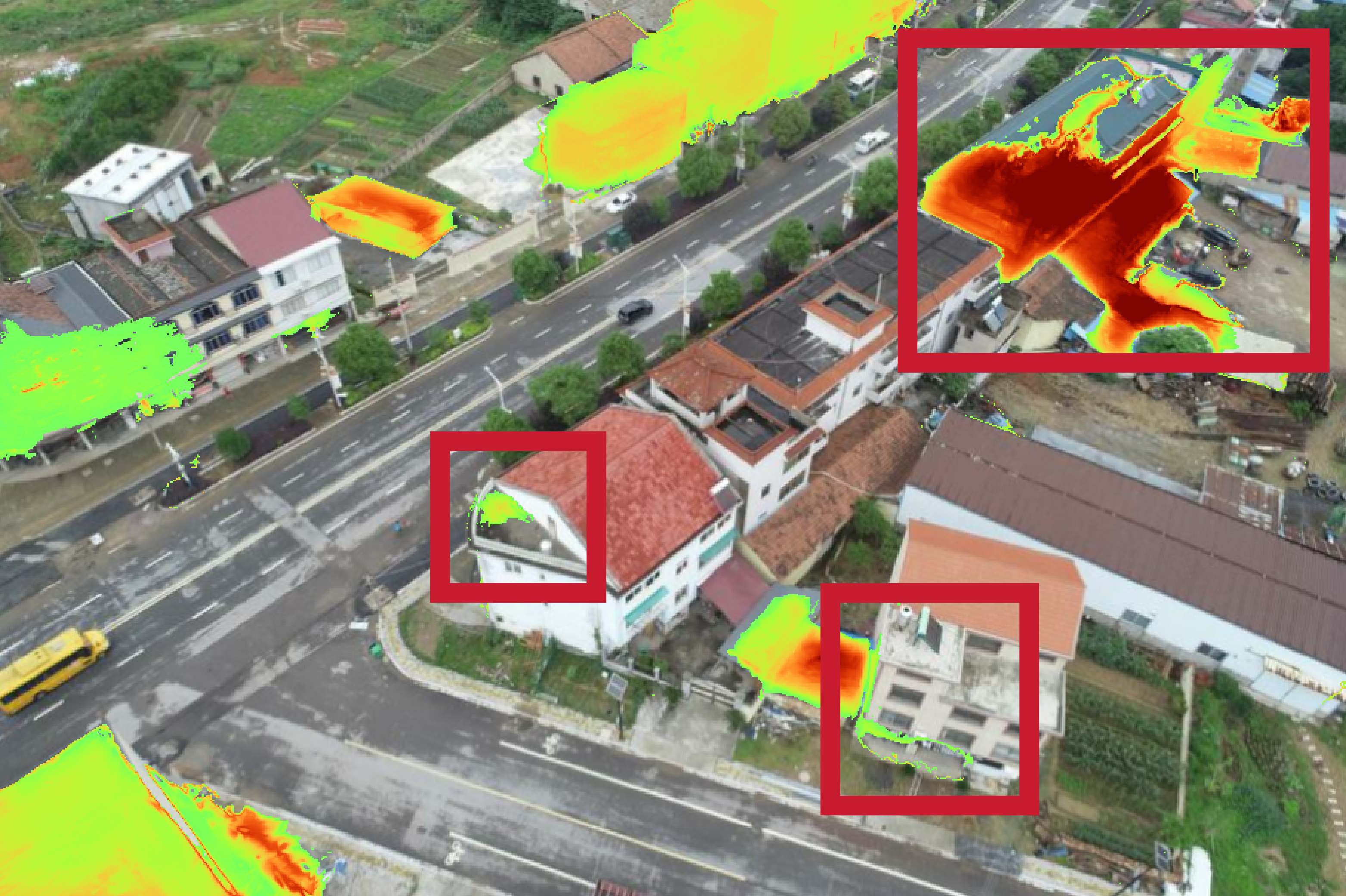} &
        \includegraphics[width=0.32\columnwidth]{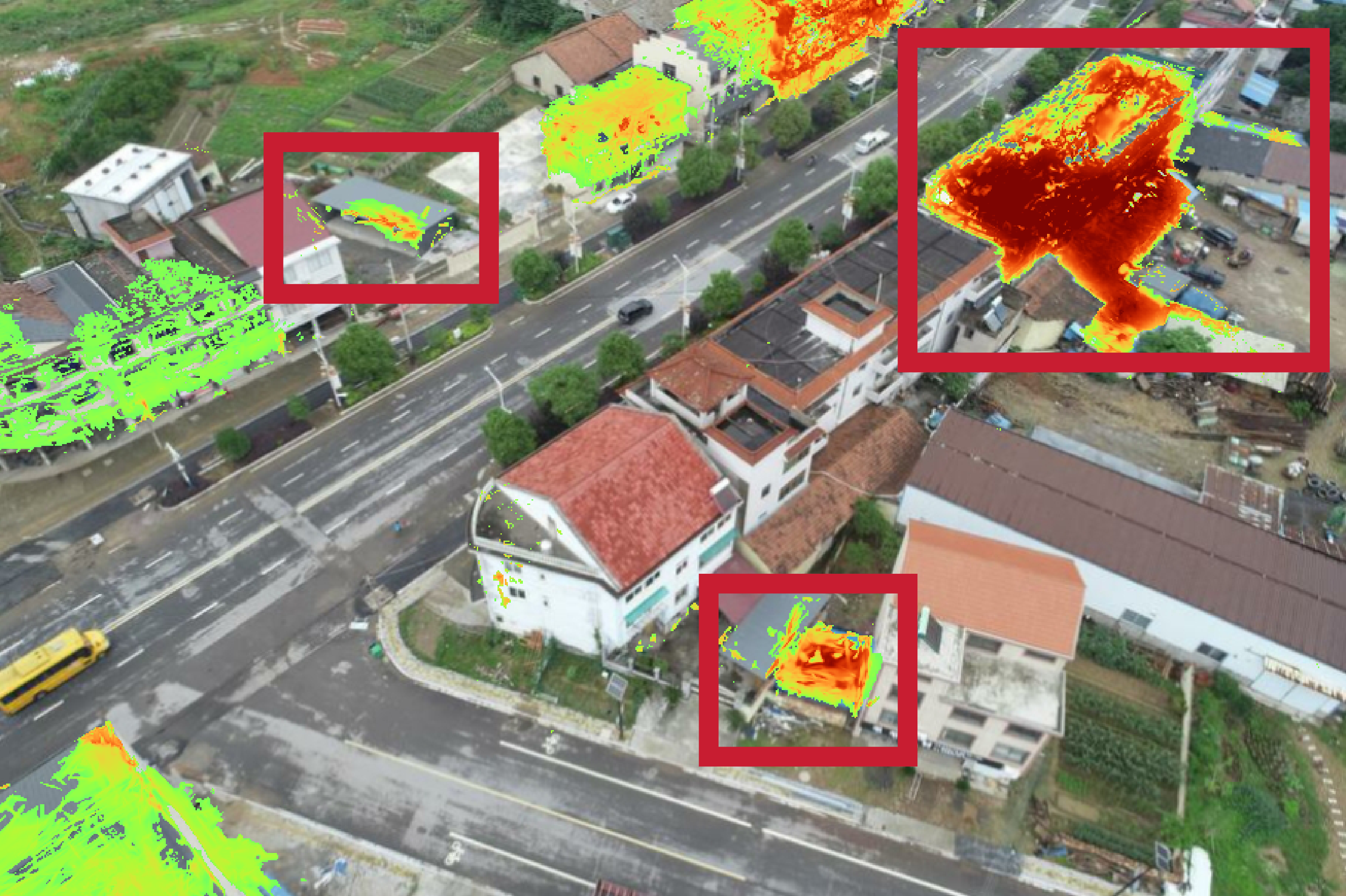} &
        \includegraphics[width=0.32\columnwidth]{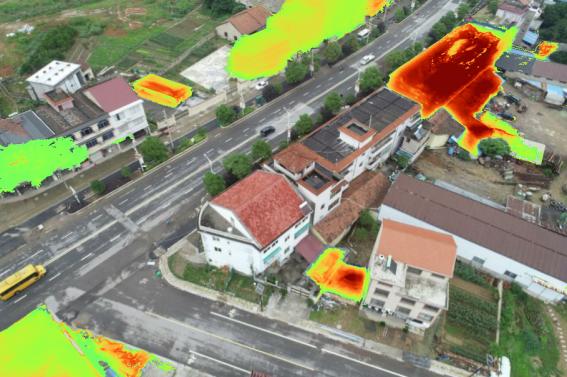} \\[-1pt]

        \textbf{(a) Projection aggregation} &
        \textbf{(b) VALA aggregation} &
        \textbf{(c) Our aggregation}
    \end{tabular}

    \caption{
    Aggregation ablation for the query ``blue roof buildings''. Projection aggregation produces incomplete and noisy responses, VALA aggregation causes local response loss, and our aggregation yields more complete results.
    }
    \label{fig:ablation-aggregation}
\end{figure}

\paragraph{Ablation on feature aggregation.}
As shown in Fig.~\ref{fig:ablation-aggregation}, (a) direct projection aggregation accumulates noisy observations caused by occlusion and boundary mixing, resulting in incomplete target responses, boundary semantic confusion, and background false activations. (b) VALA suppresses some noise but discards low-contribution yet semantically valid surface Gaussians, causing local response loss. In contrast, (c) our method combines pixel-level Gaussian contribution reliability with cross-view semantic consistency, producing more continuous, concentrated, and boundary-preserving responses. The quantitative results in Table~\ref{tab:aggregation_ablation} further validate the effectiveness of our aggregation strategy.

\subsection{Limitations and Future Works}
Our method depends on the quality of the underlying 3DGS reconstruction, as it does not explicitly optimize the reconstruction process. Reconstruction artifacts, incomplete geometry, or inaccurate Gaussian placement may affect 2D-3D feature alignment and semantic aggregation. Future work will explore joint optimization of 3D reconstruction and language-aware representations to improve small-object understanding from UAV viewpoints.

\section{Conclusion}
We propose a 3D Language Gaussian Splatting method for UAV outdoor scenes, OutLangSplat. It aligns and fuses region-based vision-language features to improve spatial consistency, reducing incomplete target responses and background misactivations.
We further introduce a training-free pixel-level contribution and consistency-aware Gaussian feature aggregation strategy to suppress unreliable responses caused by occlusions, boundary ambiguity, and noisy viewpoints. 
In addition, we construct the first accessible open-vocabulary 3D scene understanding datasets using four public UAV outdoor scenes.
Experiments demonstrate superior performance in open-vocabulary localization and semantic segmentation, while ablation studies validate the effectiveness of the proposed components.

\bibliography{references}


\end{document}